\documentclass[11pt,a4paper]{article}
\usepackage[hyperref]{acl2019}
\usepackage{times}
\usepackage{latexsym}
\usepackage{url}

\usepackage{epsfig}

\aclfinalcopy % Uncomment this line for the final submission
\title{Augmenting Data with Mixup for Sentence  Classification: An Empirical Study}
\author{
  Hongyu Guo \\
  National Research Council Canada\\
  1200 Montreal Road, Ottawa \\
  \texttt{hongyu.guo@nrc-cnrc.gc.ca} 
    \\
   \And
   Yongyi Mao \\
	Electrical Engineering \& Computer Science\\
University of Ottawa, Ottawa, Ontario\\
   \texttt{yymao@eecs.uottawa.ca} \\
   \AND
   Richong Zhang \\
   BDBC, School of Computer Science and Engineering\\
   Beihang University, Beijing, China \\
   \texttt{zhangrc@act.buaa.edu.cn} 
}

\date{}

\begin{document}
\maketitle
\begin{abstract}
Mixup~\cite{mixup17}, a  recent proposed data augmentation method through linearly interpolating inputs and modeling targets of random samples,  has demonstrated its capability of  significantly improving the predictive accuracy of the state-of-the-art   networks for image classification. However, how this technique can be applied to and what is its effectiveness on natural language processing (NLP) tasks have not been investigated. In this paper, we propose two strategies for the adaption of Mixup on sentence classification: one performs interpolation on word embeddings and another on sentence embeddings. We conduct  experiments to evaluate our methods using several benchmark datasets. Our  studies show that such  interpolation strategies serve as an  effective, domain independent data augmentation approach for sentence classification, and can result in  significant accuracy improvement for both CNN~\cite{DBLP:conf/emnlp/Kim14} and  LSTM~\cite{Hochreiter1997} models. 
\end{abstract}

\section{Introduction}
Deep learning models have achieved  state-of-the-art performance on many NLP applications, including parsing~\cite{Socher2011b}, text classification~\cite{ DBLP:conf/emnlp/Kim14,Tai2015}, and machine translation~\cite{Sutskever:2014}. 
  These models typically have millions of parameters, thus require large amounts of data for training in order for  over-fit avoidance and better model generalization. However,  collecting a large annotated data samples is time-consuming and  expensive. 
  
One  technique aiming to address such a data hungry problem is  automatic data augmentation. That is, synthetic data samples are generated as additional training data for regularizing the learning models. Data augmentation has been actively and successfully used  in computer  vision~\cite{Simard:1998:TIP:645754.668381,Krizhevsky:2017:ICD:3098997.3065386,mixup17} 
and speech recognition~\cite{Jaitly_vocaltract,DBLP:conf/interspeech/KoPPK15}. Most of these methods, however,  rely on  human knowledge for 
label-invariant data transformation, such as image scaling, flipping and rotation. Unlike in image,  there is, however, no simple rule  for label-invariant transformation in natural languages. Often, slight change of a word in a sentence can dramatically alter  the meaning of the  sentence. To this end, popular data augmentation approaches in NLP aim to transform the text with word replacements with either synonyms from handcrafted ontology (e.g.,  WordNet~\cite{Zhang:2015:CCN:2969239.2969312}) or word similarity ~\cite{DBLP:conf/emnlp/WangY15,DBLP:journals/corr/abs-1805-06201}. Such synonym-based transformation, however, can be applied to only a  portion of the vocabulary due to the fact that words having exactly or nearly the same meanings are rare.  Some other NLP data augmentation methods are often  devised for  specific domains thus makes them difficult to be applied to other domains~\cite{DBLP:conf/emnlp/SahinS18}. 

Recently, a simple yet extremely effective  augmentation method Mixup~\cite{mixup17} has been proposed and shown superior performance on enhancing the accuracy of  image classification models.   Through linearly interpolating pixels of random image pairs and their training targets, Mixup generates synthetic  examples for  training. Such training  has been shown to act as an effective model regularization strategy for image classification networks.

How Mixup can be applied to and what is its effectiveness on NLP tasks? 
We here aim to answer these questions in this paper. In specific,  we propose two strategies for the application of Mixup on sentence classification: one performs interpolation on word embedding and another on sentence embedding. We empirically show that  such  interpolation strategies serve as a simple, yet  effective data augmentation method for sentence classification, and can result in  significant accuracy improvement for both CNN~\cite{DBLP:conf/emnlp/Kim14} and  LSTM~\cite{Hochreiter1997} models. 
Promisingly, unlike traditional data augmentation in NLP, these  interpolation based  augmentation strategies are domain independent, exclusive of  human knowledge for  data transformation, and of   low additional computational cost. 

\section{Data Augmentation through Sentence Interpolation}
\subsection{MixUp for Image Classification}
 Zhang et al.~\cite{mixup17} proposed the Mixup method for image classification. 
The  idea is to generate a synthetic sample by linearly  interpolating  a pair of training samples as well as their modeling targets. 
In detail, consider a pair of samples $(x^{i}; y^{i})$ and $(x^{j} ; y^{j})$, where $x$ denotes the
input and $y$ the one-hot encoding of the corresponding class of the sample. 
The synthetic sample is generated as follows.
\begin{equation}
\widetilde{x}^{ij} = \lambda x^{i} + (1- \lambda ) x^{j}
\end{equation}
%$$\widetilde{y}_{ij} = \lambda y_{i} + (1- \lambda ) y_{j}$$
\begin{equation}
\label{equ:2}
\widetilde{y}^{ij} = \lambda y^{i} + (1- \lambda ) y^{j}
\end{equation}
where $\lambda$ is the mixing policy or mixing-ratio for the sample pair. $\lambda$ is sampled from a Beta($\alpha, \alpha$) distribution with a hyper-parameter $\alpha$. It is worth noting that, when $\alpha$  equals to one, then the Beta distribution is equivalent to an uniform distribution. 
The generated synthetic data are then fed into the model for training to minimize the loss function  such as the cross-entropy function for the supervised classification. For an efficient computation, 
the mixing happens by randomly pick one sample and then pairs it  up with another sample drawn from the same mini-batch.  

\begin{figure}[h]%[h!]
\caption{Illustration of  wordMixup (left) and senMixup (right), where the added part to the standard sentence classification model is in red rectangle.}
	\centering
\includegraphics[width=3.1in]{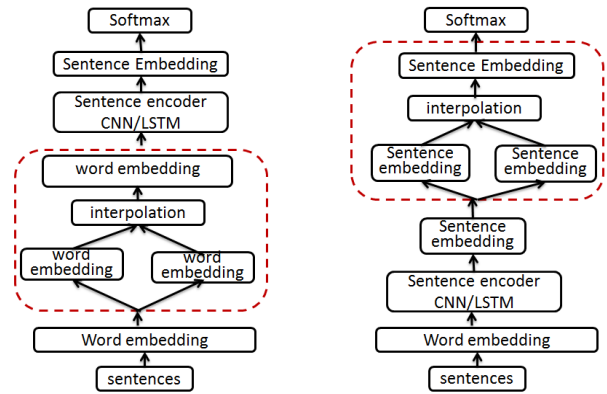}                
	\label{fig:schema}
\end{figure}

\subsection{Adaptation of Mixup for Sentence Classification}
Unlike image which is consist of pixels, sentence is composed of a sequence of words. 
Therefore, a sentence representation is often constructed to aggregate information from a sequence of words. Specifically, in a standard CNN or LSTM model, a sentence is first represented by a sequence of word embeddings, and then fed into a sentence encoder. The most popular such encoders are CNN and LSTM. The sentence embedding generated by CNN or LSTM are then passed through a soft-max layer to generate the predictive distribution over the possible target classes  for predictions. 

To this end, we propose two variants of Mixup for sentence classification. %The left subfigure depicts the wordMixup model and the right shows the schema of senMixup.  
The first one conducts sample interpolation in the word embedding space (denoted as wordMixup), and the second  on the final hidden layer of the network before it is passed to a standard soft-max layer to generate the predictive distribution over classes  (denoted as senMixup). The two models are illustrated in Figure~\ref{fig:schema}, where the standard CNN~\cite{DBLP:conf/emnlp/Kim14} or LSTM~\cite{Hochreiter1997}  model for sentence classification  corresponds to the one without the red rectangle. % Next, we will discuss them in detail.

In specific, in the wordMixup, all sentences are zero padded to the same length and then  interpolation is conducted for each dimension of each of the words in a sentence. 
 Given a piece of text, such as a sentence with $N$ words, it can be represented as a matrix $B \in R^{N \times d}$. Each row $t$ of the matrix corresponds 
to one word (denoted $B_{t}$), which is represented by a $d$-dimensional vector as provided either by a learned word embedding table or being randomly generated. 
Formally, consider a pair of samples $(B^{i}; y^{i})$ and $(B^{j} ; y^{j})$, where $B^{i}$ and $B^{j}$ denotes the embedding vectors of the input sentence pairs and  $y^{i}$ and $y^{j}$ denote the corresponding class labels of the samples using one-hot representation. Then for the $t^{th}$ word in the sentence, linear interpolation process can be formulated as: 
\begin{equation}
\widetilde{B}^{ij}_{t} = \lambda B^{i}_{t} + (1- \lambda ) B^{j}_{t} 
\end{equation}
\begin{equation}
\label{equ:2}
\widetilde{y}^{ij} = \lambda y^{i} + (1- \lambda ) y^{j}
\end{equation}

The resulting new sample ($\widetilde{B}^{ij}$;  $\widetilde{y}^{ij}$) is then used for training.  

In  senMixup, the hidden embeddings (with the same   dimension) for the two sentences are first generated by an encoder such as CNN or LSTM. Next, the pair of sentence embeddings are interpolated linearly. In specific, let $f$ denote the sentence encoder, then a pair of sentences  $B^{i}$ and $B^{j}$ will be first encoded into a pair of sentence embeddings $f(B^{i})$ and $f(B^{j})$, respectively. In this case, the mixing is conducted for each $k^{th}$ dimension of the sentence embedding, as follows. 
\begin{equation}
\widetilde{B}^{ij}_{\{k\}} = f(B^{i})_{\{k\}} + (1- \lambda ) f(B^{j})_{\{k\}}
\end{equation}
%$$\widetilde{y}^{ij} = y^{i} + (1- \lambda ) y_{j}$$
\begin{equation}
\label{equ:2}
\widetilde{y}^{ij} = \lambda y^{i} + (1- \lambda ) y^{j}
\end{equation}
Finally, the embedding vector $\widetilde{B}^{ij}$ will be passed to a softmax layer to  produce a distribution over the possible target classes. For training, we use multi-class cross entropy loss.

\section{Experiment}
%\subsection{Data Sets and Baselines}
We evaluate the proposed methods with five benchmark tasks for sentence classifications. 
\begin{itemize}
	\item \textbf{TREC} is a question dataset to categorize a question into six question types~\cite{Li:2002:LQC:1072228.1072378}. 
	\item   \textbf{MR} is a movie review dataset for detecting positive/negative reviews~\cite{Pang2005}. 
	\item \textbf{SST-1} is the Stanford Sentiment Treebank with five categories labels~\cite{Socher2013}. 
	\item \textbf{SST-2} dataset is the same as SST-1 but with neutral reviews
removed and binary labels.
	\item \textbf{Subj} is a subjectivity detection dataset for classifying a sentence as being subjective or
objective~\cite{DBLP:conf/acl/PangL04}. 
\end{itemize}
 
The summary of the data sets  is presented in Table~\ref{tab:data}. %~\cite{DBLP:conf/emnlp/Kim14}. 
Note that, for comparison purposes on the SST-1 and SST-2 datasets, following~\cite{DBLP:conf/emnlp/Kim14,Tai2015}, we trained the models using both phrases and sentences, but only evaluated sentences at test time.

We evaluate our wordMixup and senMixup  using both CNN and LSTM for sentence classification. We implement the  CNN model  exactly as reported in~\cite{DBLP:conf/emnlp/Kim14,kim/code}. For LSTM, we just simply replace the convolution/pooling components in CNN with standard LSTM units as implemented in~\cite{Abadi:2016:TSL:3026877.3026899}. The final feature map of CNN and the final state of LSTM are passed to a logistic regression classifier for label prediction.

To evaluate our models in terms of their regularization effects on the training, we present four word embedding settings: random and  trainable word embedding (denoted \textbf{RandomTune}),  random and fix word embedding (denoted \textbf{RandomFix}), pre-trained and tunable word embedding  (denoted \textbf{PretrainTune}), and pre-trained fix word embedding (denoted \textbf{PretrainFix}).

%\iffalse
\begin{table}[h!]
  \centering
\begin{tabular}{l|c|c|c|c|c}\hline
Data& c& l &N &V &Test\\ \hline
TREC &6 &10& 5952 &9592 &500\\
SST-1 &5 &18 &11855 &17836  &2210\\
SST-2 &2 &19 &9613 &16185  &1821\\
Subj &2 &23 &10000 &21323& CV\\
MR& 2 &20& 10662 &18765  &CV\\
	\hline
\end{tabular}
  \caption{ Summary  for the datasets after tokenization.
c: number of target labels. l: average sentence length. 
N: number of samples. V: vocabulary size. Test:
test set size (CV means  no standard train/test split was provided 
and thus 10-fold CV was used).}
  \label{tab:data}
\end{table}
%\fi

\begin{table*}[h!]
  \centering
%		\scalebox{0.88}{
\begin{tabular}{l|c|c|c|c|c}\hline
\textbf{RandomTune} &Trec& SST-1& SST-2&Subj  & MR\\ \hline \hline
CNN- KIM Impl.~\cite{DBLP:conf/emnlp/Kim14} &91.2&45.0&82.7&89.6&76.1\\
CNN- HarvardNLP Impl.~\footnote{https://github.com/harvardnlp/sent-conv-torch}&88.2&42.2&83.5&89.2&75.9\\
CNN - Our Impl. &90.2$\pm$0.20&43.6$\pm$0.19&82.3$\pm$0.47&90.6$\pm$0.45&75.5$\pm$0.36\\ 
\hline
CNN+wordMixup&90.9$\pm$0.42&45.2$\pm$0.90&82.8$\pm$0.45&\textbf{92.9$\pm$0.41}&\textbf{78.0$\pm$0.39}\\
  CNN+senMixup&\textbf{92.1$\pm$0.31}&\textbf{45.2$\pm$0.22}&\textbf{83.0$\pm$0.35}&92.7$\pm$0.38&77.9$\pm$0.76\\
	\hline
\end{tabular}
%}
  \caption{Accuracy (\%) of the testing methods using CNN (with randomly initialized, trainable embeddings).  We report mean scores over 10 runs with standard deviations (denoted $\pm$). Best results highlighted in Bold.}
  \label{tab:accuracy:cnn:randomtune}
\end{table*}

\begin{table*}[h!]
  \centering
		\scalebox{0.9}{
\begin{tabular}{l|c|c|c|c|c}\hline
\textbf{RandomTune} &Trec& SST-1& SST-2&Subj  & MR\\ \hline \hline
LSTM-StanfordNLP Impl.~\cite{Tai2015} &N/A&46.4&84.9&N/A&N/A\\
LSTM-AgrLearn Impl.~\cite{DBLP:journals/corr/abs-1807-10251} &N/A&N/A&N/A&90.2&76.2\\
LSTM - Our Impl. &86.5$\pm$0.61&45.9$\pm$0.58&84.4$\pm$0.35&90.9$\pm$0.42&77.2$\pm$0.75\\
\hline
LSTM +   wordMixup&\textbf{90.5$\pm$0.50}&48.2$\pm$0.18&86.3$\pm$0.35&\textbf{93.1$\pm$0.49}&\textbf{78.0$\pm$0.33}\\
LSTM +   senMixup&89.4$\pm$0.40&\textbf{48.3$\pm$0.77}&\textbf{86.7$\pm$0.33}&91.9$\pm$0.34&77.9$\pm$0.33\\ \hline
\end{tabular}
}
  \caption{Accuracy (\%) obtained by the testing methods using LSTM (with randomly initialized, trainable embeddings).  We report mean scores over 10 runs with standard deviations  (denoted $\pm$). Best results highlighted in Bold.}
  \label{tab:accuracy:lstm}
\end{table*}

\begin{figure*}[h]
    \centering
    \includegraphics[height=3.3in]		 {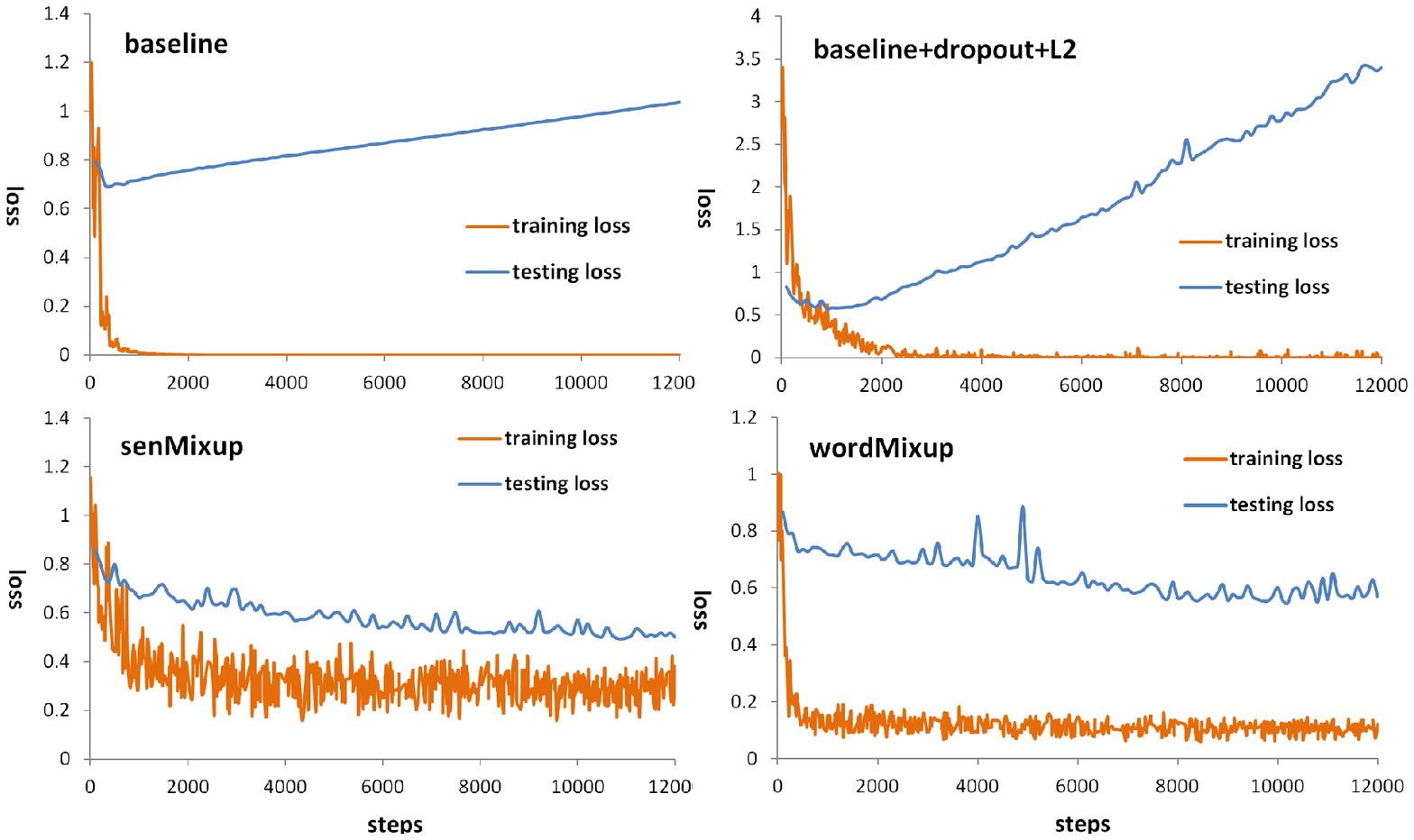}
        \label{fig:conv}
				\caption{Training and testing entropy loss obtained by the baseline CNN without dropout (top-left), baseline CNN with dropout and L2 (top-right), wordMixup (bottom-right), and senMixup (bottom-left).}
\end{figure*}%

%\subsection{Hyperparameters and Training Details}
In our experiments, following the exact implementation and settings in~\cite{kim/code} we use  filter sizes of 3, 4, and 5, each with 100 feature maps; dropout rate of 0.5 and L2 regularization of 0.2 for the baseline CNN and LSTM.  For datasets without a standard dev set we
randomly select 10\% of training data as dev set. Training is done through Adam~\cite{DBLP:journals/corr/KingmaB14} over   mini-batches of size 50.
The pre-trained word embeddings are 300 dimensional GloVe~\cite{Pennington14glove:global}. The hidden state dimension  for LSTM is 100. 

For senMixup and wordMixup, the mixing policy $\alpha$ is set to the default value of one. %, meaning that the interpolation is conducted based on a uniformly sampled random ratio between 0.0 and 1.0. 
Also, following the original Mixup~\cite{mixup17}, we did not use dropout or L2 constraint for the wordMixup and senMixup models. 

We train each model 10 times each with 20000 steps, and compute their mean test errors and standard deviations.

%\subsection{Results for RandomTune}
\subsection{Main Results}
RandomTune has the largest number of parameters,  compared to RandomFix, PretrainTune, and PretrainFix, and thus requires  a strong regularization method to avoid  over-fit the training data. We, therefore, focus our experiments on the RandomTune setting. 
The  results on  the RandomTune setting are presented in Table~\ref{tab:accuracy:cnn:randomtune}. 

The results in Table~\ref{tab:accuracy:cnn:randomtune} show that wordMixup and senMixup  provide good regularization to  CNN, resulting in  accuracy improvement on all the five  testing datasets. For example, in SST-1 and MR, the relative improvement was over 3.3\%. 
Interestingly, both wordMixup and senMixup failed  to significantly improved over the baseline against the SST-2 dataset; with senMixup slightly outperformed the baseline with only 0.7\%. 
Also, results in Table~\ref{tab:accuracy:cnn:randomtune}  suggest that senMixup and wordMixup were quite competitively, in terms of predicticve performance obtained, against the five testing datasets. For example, on the Trec dataset, senMixup outperformed senMixup with 1.2\%, but for the other four datasets, the two methods obtained very similiar predictive accuracy.

\noindent \textbf{Regularization Effect} We  plot the training and testing cross-entropy loss across the first 12K training steps  on the MR dataset in Figure~\ref{fig:schema}.   
Figure~\ref{fig:schema} shows that with (top-left subfigure) or without (top-right subfigure) dropout, the training loss of CNN drops to zero quickly and provides no training signal for further tuning the networks. On the otherhand, the training loss of wordMixup  (bottom-right subfigure) keeps above  zero during the training, continuously provide training signal for the network learning.  Also, the  training loss curve of senMixup (bottom-left subfigure) maintains a relatively high level, allowing the model to keep tuning. The relatively higher training loss of both wordMixup and senMixup is due to the much larger space of the mixed samples, thus  preventing the model from being over-fitted by limited number of individual examples

\noindent \textbf{LSTM Networks as Sentence Encoder} We also evaluate the effect of using LSTM as the sentence encoder. 
Results in Table~\ref{tab:accuracy:lstm} show that, similar to the case of using CNN as sentence encoder, wordMixup and senMixup with LSTM as encoder  also improved the predictive performance of the baseline models. For example,  the largest improvements came from the Trec and SST-1 cases (with relative improvement of 4.62\% and 5.22\%), which have six and five classes, respectively. Results in the table also suggest that, on the Subj dataset, wordMixup outperformed senMixup with 1.2\%, but for the other four datasets, the two methods performed comparably well. 

One notable fact when compared with the CNN-based models as presented in Table~\ref{tab:accuracy:cnn:randomtune} is that, against the SST-2 data sets, both wordMixup and senMixup with LSTM here were able to improve over the baseline with about 2\%.

\begin{table*}[h!]
  \centering
	%	\scalebox{0.86}{
\begin{tabular}{l|c|c|c|c|c}\hline
\textbf{RandomFix} & Trec&SST-1& SST-2&Subj  & MR\\ \hline 
CNN &88.4$\pm$0.52&40.3$\pm$0.77&\textbf{80.4$\pm$0.17}&88.2$\pm$0.50&72.9$\pm$0.74\\
CNN + wordMixup&\textbf{90.9$\pm$0.58}&40.5$\pm$1.17&77.5$\pm$0.33&89.3$\pm$0.47&\textbf{74.2$\pm$1.15}\\
  CNN +   senMixup&88.8$\pm$1.10&\textbf{41.0$\pm$0.64}&77.6$\pm$0.76&\textbf{90.5$\pm$0.36}&72.6$\pm$0.67\\
	\hline
\end{tabular}
%}
  \caption{Accuracy (\%) obtained by the testing methods using CNN with randomly initialized and fixed embeddings. Best results highlighted in Bold.}
		\label{tab:accuracy:cnn:randomfix}
\end{table*}

\begin{table*}[h!]
  \centering
		%\scalebox{0.86}{
\begin{tabular}{l|c|c|c|c|c}\hline
\textbf{PretrainTune}&Trec & SST-1& SST-2&Subj  & MR\\ \hline %\hline
CNN &92.1$\pm$0.12&46.3$\pm$0.35&86.9$\pm$0.49&94.4$\pm$0.36&79.8$\pm$0.60\\
CNN + wordMixup&93.7$\pm$0.80&48.2$\pm$0.91&87.1$\pm$0.26&94.7$\pm$0.45&\textbf{81.3$\pm$0.28}\\
  CNN +   senMixup&\textbf{93.3$\pm$0.23}&\textbf{48.6$\pm$0.23}&\textbf{87.2$\pm$0.35}&\textbf{94.9$\pm$0.34}&80.6$\pm$0.56\\
	\hline
\end{tabular}
%}
  \caption{Accuracy (\%) of the testing methods using CNN with pre-trained GloVe and trainable embeddings. Best results highlighted in Bold.}
  \label{tab:accuracy:cnn:pretraintune}
\end{table*}

\begin{table*}[h!]
  \centering
		%\scalebox{0.86}{
\begin{tabular}{l|c|c|c|c|c}\hline
\textbf{PretrainFix} &Trec& SST-1& SST-2&Subj  & MR\\ \hline 
CNN &92.0$\pm$0.2&44.6$\pm$0.56&\textbf{85.7$\pm$0.33}&94.5$\pm$0.36
&79.7$\pm$0.68\\
CNN + wordMixup&94.2$\pm$0.52& \textbf{46.6$\pm$0.85}&84.5$\pm$0.54&94.3$\pm$0.23 &79.7$\pm$0.52\\

  CNN +   senMixup&\textbf{94.8$\pm$0.35}&46.5$\pm$0.23&84.7$\pm$0.48&\textbf{95.0$\pm$0.22}
&\textbf{80.3$\pm$0.57}\\
\hline
\end{tabular}
%}
   \caption{Accuracy (\%) obtained by the testing methods using CNN with pretrained GloVe and fixed embeddings. Best results highlighted in Bold.} 
		\label{tab:accuracy:cnn:pretrainfix}
\end{table*}

%\subsection
\noindent \textbf{Results for RandomFix, PretrainTune, and PretrainFix}
Results for the settings of RandomFix, PretrainTune, and PretrainFix are presented in Tables~\ref{tab:accuracy:cnn:randomfix},~\ref{tab:accuracy:cnn:pretraintune}, and ~\ref{tab:accuracy:cnn:pretrainfix}, respectively. Results in these three tables further confirm that  data augmentation through wordMixup and senMixup can improve the predictive performance of the base models, except on the SST-2 dataset.  
On the SST-2 data set, both wordMixup and senMixup degraded the predictive accuracy  of the baseline when the word embeddings were not allowed to be tuned during training. With  learnable word embeddings, although both wordMixup and senMixup failed to significantly improve over the baseline on this dataset, but they did obtain  similar predictive performance as the baseline. 
In short, our experiments here suggest that when the word embeddings are tuned, both wordMixup and senMixup are able to improve the predictive accuracy of the base models.

\section{Conclusion and Future Work}
Inspired by the success of Mixup, a  simple and effective data augmentation method through sample interpolation for image recognition, we  investigated two variants of Mixup for sentence classification. We empirically show that they can improve the accuracy upon  both CNN and LSTM sentence classification models. Interestingly,  our  studies here show that such  interpolation strategies can serve as an  effective, domain independent regularizer for overfitting avoidance  for sentence classification.

We plan to investigate some lately proposed variants of Mixup, such as Manifold Mixup~\cite{DBLP:journals/corr/abs-1806-05236}, where interpolation is performed in a randomly selected  layer of the networks, and AdaMixup~\cite{DBLP:journals/corr/abs-1809-02499}, which addresses the manifold intrusion issues in Mixup. We are also interested in questions such as what the mixed sentences look like and why interpolation works for sentence classification.

\bibliographystyle{acl_natbib}
\bibliography{acl2019}

\end{document}